\documentclass[sigconf,natbib=true,anonymous=False]{acmart}

\usepackage{balance}

\newcommand{\circled}[1]{\resizebox{0.95em}{!}{\textcircled{\scriptsize#1}}}

\AtBeginDocument{%
  \providecommand\BibTeX{{%
    \normalfont B\kern-0.5em{\scshape i\kern-0.25em b}\kern-0.8em\TeX}}}

\copyrightyear{2024}
\acmYear{2024}
\setcopyright{acmlicensed}\acmConference[SIGIR '24]{Proceedings of the 47th International ACM SIGIR Conference on Research and Development in Information Retrieval}{July 14--18, 2024}{Washington, DC, USA}
\acmBooktitle{Proceedings of the 47th International ACM SIGIR Conference on Research and Development in Information Retrieval (SIGIR '24), July 14--18,
2024, Washington, DC, USA}
\acmDOI{10.1145/3626772.3657960}
\acmISBN{979-8-4007-0431-4/24/07}

\acmConference[SIGIR '24]{the 47th International ACM SIGIR Conference
on Research and Development in Information Retrieval}{July 14--18,
  2024}{Washington D.C., USA}

\acmSubmissionID{8032}

\begin{document}

\title{Enhancing Criminal Case Matching through Diverse Legal Factors}

\author{Jie Zhao}
\affiliation{
  \institution{Xidian University}
  \city{Xi'an}
  \country{China}}
\email{jzhao1992@stu.xidian.edu.cn}
\author{Ziyu Guan}
\authornote{Corresponding author.}
\affiliation{
  \institution{Xidian University}
  \city{Xi'an}
  \country{China}}
\email{zyguan@xidian.edu.cn}
\author{Wei Zhao}
\affiliation{
  \institution{Xidian University}
  \city{Xi'an}
  \country{China}}
\email{ywzhao@mail.xidian.edu.cn}
\author{Yue Jiang}
\affiliation{
  \institution{Xidian University}
  \city{Xi'an}
  \country{China}}
\email{22031212489@stu.xidian.edu.cn}

\renewcommand{\shortauthors}{Jie Zhao, Ziyu Guan, Wei Zhao, \& Yue Jiang}

\begin{abstract}
Criminal case matching endeavors to determine the relevance between different criminal cases. Conventional methods predict the relevance solely based on instance-level semantic features and neglect the diverse legal factors (LFs), which are associated with diverse court judgments. Consequently, comprehensively representing a criminal case remains a challenge for these approaches. Moreover, extracting and utilizing these LFs for criminal case matching face two challenges: (1) the manual annotations of LFs rely heavily on specialized legal knowledge; (2) overlaps among LFs may potentially harm the model's performance. In this paper, we propose a two-stage framework named Diverse Legal Factor-enhanced Criminal Case Matching (DLF-CCM). Firstly, DLF-CCM employs a multi-task learning framework to pre-train an LF extraction network on a large-scale legal judgment prediction dataset. In stage two, DLF-CCM introduces an LF de-redundancy module to learn shared LF and exclusive LFs. Moreover, an entropy-weighted fusion strategy is introduced to dynamically fuse the multiple relevance generated by all LFs. Experimental results validate the effectiveness of DLF-CCM and show its significant improvements over competitive baselines. Code: \url{https://github.com/jiezhao6/DLF-CCM.}

\end{abstract}



\maketitle

\section{Introduction}

Legal case matching aims to determine the relevance between different judicial cases. It plays a crucial role in enhancing the effectiveness of intelligent legal systems, especially for legal case retrieval. 
Legal cases are generally categorized into civil, criminal, and administrative cases \cite{legal_feature}. 
Among them, criminal cases are closely tied to the freedoms and even the lives of individuals, necessitating a higher level of accuracy for matching models. In this work, we focus on the task of criminal case matching (CCM).

In addressing this task, some studies have considered cases as general long-form text documents \cite{bert-PLI,lawformer}. 
Other works investigate deeply into the exploration of domain knowledge within legal texts \cite{legal_feature}, such as utilizing logical relationships between different sections of semi-structured legal texts \cite{structure-aware} and introducing the knowledge of legal articles \cite{law-match}. 
Nevertheless, these approaches simply rely on instance-level semantic representations for predicting, neglecting the diverse legal factors (LFs) introduced below.

\begin{figure}
    \centering
    \includegraphics[width=0.35\textwidth,height=0.1\textwidth]{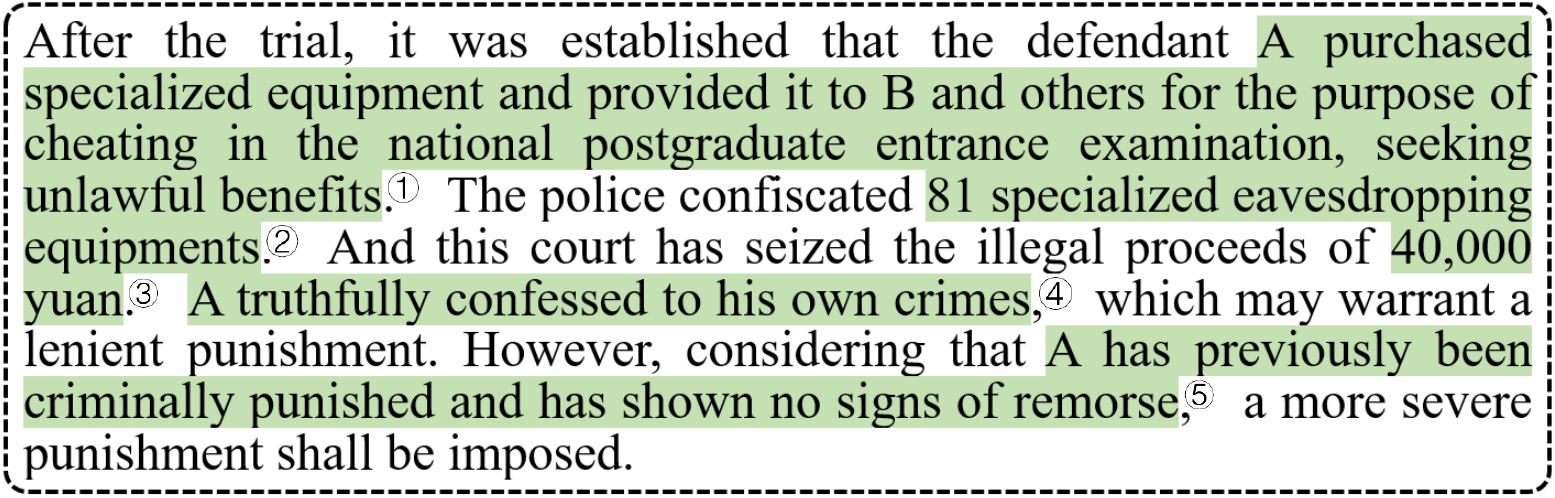}
    
    \caption{A criminal case (translated) with diverse LFs. The highlighted segments \circled{1}, \circled{4}, and \circled{5} represent ARF; the \circled{1}, \circled{2}, and \circled{3} represent CRF; the \circled{2}, \circled{3}, \circled{4} and \circled{5} represent TRF.}

    \Description{An example of a criminal case with diverse legal factors.}
    \label{fig:example}
\end{figure}

In criminal cases, the factual descriptions play crucial roles in supporting the court judgments, including applicable legal articles, charges, and prison terms. \citeauthor{multi-law-LJP}~\cite{multi-law-LJP} have observed that the importance of different components within cases varies across these distinct judgments. However, their work focuses on distinguishing charge- and term-related law articles to enhance legal judgment prediction (LJP), rather than case matching. 
In our study, we further categorize the factual details of criminal cases into article-related factor (ARF), charge-related factor (CRF), and term-related factor (TRF). 
In general, ARF involves the objective actions and subjective attitudes of suspects; CRF involves the instances of constitutive elements \cite{cecp} of a charge; and TRF concerns the crime severity and the suspects' attitudes towards confession.
Figure~\ref{fig:example} shows a criminal case and we highlight several segments exhibiting clear correlations with different LFs. 
Conventional methods, due to their reliance on a single case representation, face challenges in capturing various LFs, thus failing to learn comprehensive case representations.

We argue that extracting and utilizing these LFs can enhance the performance of CCM. The underlying assumption is that similar cases should yield similar judgmental results, supported by similar LFs. 
However, extracting and utilizing LFs face two challenges: (1) There is no annotated data for LFs. The manual annotation of LFs relies heavily on specialized legal knowledge and is highly expensive; (2) As shown in Figure~\ref{fig:example}, there are overlaps among LFs. Directly utilizing LFs may potentially harm the matching performance \cite{redundancy}.

To address these issues, we propose a two-stage framework for CCM, named Diverse Legal Factor-enhanced Criminal Case Matching (DLF-CCM). 
Considering the crucial role of different LFs in supporting various court decisions, DLF-CCM initially pre-trains an LF extraction network driven by the LJP task to address the challenge (1). The pre-training objective is formulated to enable a model to accurately predict judgmental results based on extracted LFs. 
Subsequently, DLF-CCM retains the LF extractor and introduces an LF de-redundancy module to learn a shared LF and exclusive ARF, CRF, and TRF, to address the challenge (2).
All the learned LFs are utilized to predict the relevance of two criminal cases.
Moreover, based on the confidence of relevance prediction of each LF, we introduce an entropy-weighted multi-relevance fusion module to aggregate the predictions of all LFs.

To our knowledge, we first analyze and exploit diverse LFs in CCM. A novel two-stage matching framework is proposed to avoid the laborious labeling of LFs and eliminate the redundancy among LFs. Experimental results verify the effectiveness of DLF-CCM and show its significant improvements over competitive baselines.

\section{Related Work}
Inchoate works for CCM mainly relied on manual knowledge or feature engineering \cite{survey}, severely limiting their applicability across diverse scenarios. Recently, researchers have turned to taking advantage of deep learning and proposed a series of network-based and text-based methods \cite{law-match}. 
The former approaches are tailored for common law systems and utilize inter-citation links, including links between cases and status, cases and cases, and status and status to form a citation network \cite{Hier-SPCNet-1,Hier-SPCNet-2}. 
The latter methods compute the semantic similarity between cases.
\citeauthor{bert-PLI}~\cite{bert-PLI} employed BERT to capture paragraph-level semantic representations.
\citeauthor{lawformer}~\cite{lawformer} pre-trained the Longformer \cite{longformer} on a large-scale legal corpus for long legal documents understanding. \citeauthor{iot-match}~\cite{iot-match} extracted rationales and generated explanations for explainable legal case matching. 
\citeauthor{law-match}~\cite{law-match} utilized the mediation effect of law articles and the direct effect of key circumstances in cases. \citeauthor{structure-aware}~\cite{structure-aware} employed
an asymmetric encoder-decoder architecture to utilize the structural information contained in legal case documents. \citeauthor{prompt}~\cite{prompt} utilized prompt learning to encode legal facts and issues.

Another research area related to our work is legal judgment prediction (LJP). LJP aims to predict probable court judgments based on factual descriptions of cases. Most LJP methods conceptualize the task within the framework of multi-task learning. 
Some studies pay attention to the parameter sharing among subtasks \cite{LJP-relational}. For example, \citeauthor{ladan}~\cite{ladan} proposed a novel graph neural network to learn subtle differences among confusing law articles, thereby enhancing the fact representations.
Other works aim to mutually enhance one another by capturing the interdependencies among three legal subtasks. \citeauthor{topological}~\cite{topological} utilized a directed acyclic graph to capture the topological dependencies among subtasks. \citeauthor{bi-feedback}~\cite{bi-feedback} designed a backward verification mechanism to leverage the dependencies of prediction results. \citeauthor{multi-law-LJP}~\cite{multi-law-LJP} introduced distinct role embeddings for charge- and prison term-related law articles to model the dependencies between different types of articles and terms.

\section{Methodology}

\subsection{Problem Formulation}
Typically, a criminal case $x$ is characterized by a sequence of words describing its fact. Let $\mathcal{D} = \{ (x_s, x_t, y)_i \}_{i=1}^{\lvert \mathcal{D} \rvert}$ be a set of labeled data tuples, where $x_s$ and $x_t$ are the source and target case respectively, and $y \in \mathcal{Y}$ is the human-annotated matching label (e.g., 0/1/2 for mismatch/partially match/match). The task of criminal case matching is to develop a model $x_s \times x_t \rightarrow \mathcal{Y}$ based on $\mathcal{D}$.

Furthermore, we employ the LJP task to assist in the extraction of LFs. Given a LJP dataset $\mathcal{D}_{ljp} = \{ (x_{ljp}, y_1, y_2, y_3)_i \}_{i=1}^{\lvert \mathcal{D}_{ljp} \rvert}$, where $x_{ljp}$ is the criminal case, and $y_1/y_2/y_3 \in \mathcal{Y}_1/\mathcal{Y}_2/\mathcal{Y}_3$ are the label of article/charge/term respectively, the task of LJP becomes learning a model that can accurately predict $(y_1, y_2, y_3)$ for a given $x_{ljp}$.
In this paper, we use bold face lower/upper case letters to denote vectors/matrices respectively.

\subsection{The Proposed DLF-CCM}
\begin{figure}
    \centering
    \includegraphics[width=0.4\textwidth]{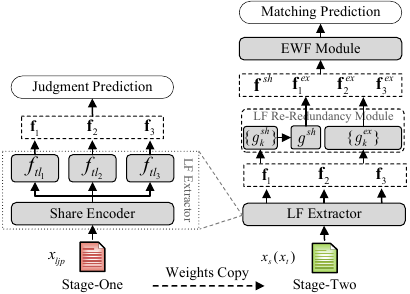}
    
    \caption{The framework of DLF-CCM.}
    \Description{The framework of DLF-CCM.}
    \label{fig:framework}
\end{figure}

In DLF-CCM, a separate LJP task is introduced to support the extractions of ARF, CRF, and TRF. We employ a simple but effective parameter-sharing multi-task learning framework to pre-train an LF extractor. 
Subsequently, the designed matching network is composed of an LF de-redundancy (LFDR) module and an entropy-weighted multi-relevance fusion (EWF) module. The LFDR module learns a shared LF and three exclusive LFs from the extracted LFs, thereby eliminating redundant information among them. Based on the confidence of matching prediction results from different LFs, the DWF module dynamically fuses these predictions and produces the final prediction.
The proposed DLF-CCM is illustrated in Figure~\ref{fig:framework}.

\textbf{Judgment-Driven Pre-Training:}
The case $x_{ljp}$ is initially encoded using a Transformer-based encoder $f_{enc}(\cdot)$ such as BERT \cite{bert}. Subsequently, we add an individual Transformer layer $f_{tl_{k}}(\cdot)$ for each subtask of LJP to extract subtask-dependent LFs, i.e., ARF, CRF, and TRF, where $k \in \{1,2,3\}$ represents the subtask index. 
The embeddings of the [CLS] token are then utilized as representations for each LF, and the corresponding subtask label is predicted by the individual classification layer $f_{cls_{k}}(\cdot)$. We employ the cross-entropy loss for pre-training:
\begin{equation}
\label{eq:loss_pretrian}
    \mathcal{L}_{pt} = - \frac{1}{k}
   \sum\nolimits_{k=1}^{3} \sum\nolimits_{j=1}^{\lvert \mathcal{Y}_k \rvert} 
    \mathbf{y}_{k, j} \log (f_{cls_k}(f_{tl_{k}}(f_{enc}(x_{ljp})))_{k, j}),
\end{equation}
where $\mathbf{y}_{k}$ is the ground-truth label of subtask $k$.

\textbf{LF De-Redundancy:} After pre-training, we drop the subtask-specific classification layer and retain the networks $f_{enc}(\cdot)$ and $f_{tk_k}(\cdot)$ as the LF extractor. Thus we can obtain three LFs $\mathbf{f}_1$, $\mathbf{f}_2$, and $\mathbf{f}_3$ for a criminal case, corresponding to ARF, CRF, and TRF, respectively. 
Throughout the pre-training process, each LF is dedicated to minimizing the error rate of the respective subtask. Consequently, a notable concern is the potential redundancy among these LFs, which may degrade the subsequent matching process \cite{redundancy}. 
Drawing on this, we propose learning exclusive LFs, which do not contain redundant information, and a shared LF.

We introduce three affine transformations (ATs) $g_k^{ex}(\cdot)$, each dedicated to learning one of the three exclusive LFs, i.e., $\mathbf{f}_k^{ex} = g_k^{ex}(\mathbf{f}_k)$. For the shared LF learning, we propose a hierarchical AT network. First, intermediate shared LFs are learned based on two LFs, expressed as $\mathbf{f}_k^{sh} = g_k^{sh} ([\mathbf{f}_k; \mathbf{f}_{1 + [k~\text{mod}~3]}])$, where ``$;$'' denotes the concatenation operation and $g_k^{sh}(\cdot)$ is an AT. Then, the final shared LF is the AT of $\mathbf{f}_k^{sh}$, i.e., $\mathbf{f}^{sh} = g^{sh} ([\mathbf{f}_1^{sh}; \mathbf{f}_2^{sh}; \mathbf{f}_3^{sh}])$. 
To supervise the learning of these LFs, we introduce a discriminator $D$, trained with the cross-entropy loss on exclusive LFs, to identify different types of LFs. The training of $D$ and the entire matching network are carried out in an alternating manner. 
Based on the discriminator $D$, we introduce the following loss for exclusive LFs learning: 
\begin{equation}
\label{eq:loss_specific}
    \mathcal{L}_{ex} = - \mathbb{E}_{\mathbf{f}_k^{ex} \sim g_k^{ex}} [ \log P_{D} (k \mid \mathbf{f}_k^{ex}) ].
\end{equation}

On the other hand, we expect the shared $\mathbf{f}_k^{sh}$ and $\mathbf{f}^{sh}$ to be agnostic of LF type, and we derive the following entropy-based loss\footnote{$\mathcal{L}_{ex}$ and $\mathcal{L}_{sh}$ are employed for both source and target cases.}:
\begin{equation}
\label{eq:loss_share}
    \mathcal{L}_{sh} = - \frac{1}{4} \sum\nolimits_{k=0}^{3} \mathcal{H} (P_{D} (\mathbf{f}_k^{sh})).
\end{equation}
where $\mathcal{H}(\cdot)$ represents the entropy of a probability distribution. When $k = 0$, $\mathbf{f}_0^{sh} = \mathbf{f}^{sh}$.

\textbf{Entropy-Weighted Multi-Relevance Fusion:} We posit that two similar cases should exhibit similarity in various LFs. Therefore, we predict the relevance of two cases based on each LF (i.e., shared LF and exclusive LFs) and subsequently fuse the relevance to collaboratively predict the final matching label.

Specifically, we add a classifier $g_{cls_{k}}(\cdot)$ for each LF\footnote{Here, $k \in \{0,1,2,3\}$, with $g_{cls_{0}}$ for $\mathbf{f}^{sh}$ and $g_{cls_{1/2/3}}$ for $\mathbf{f}^{ex}_{1/2/3}$.}.
The input to $g_{cls_{k}}$ is $[\mathbf{f}_{x_s}; \mathbf{f}_{x_t}; \mathbf{f}_{x_s} \oplus \mathbf{f}_{x_t}; \mathbf{f}_{x_s} \odot \mathbf{f}_{x_t}]$, where $\mathbf{f}_{x_s}$ and $\mathbf{f}_{x_t}$ can be any one of the four LFs, and $\oplus$/$\odot$ denote element-wise addition/multiplication.
Let $\mathbf{z}_{k}$ be the logits produced by the final layer of $g_{cls_k}(\cdot)$, and $\mathcal{H}_{k}$ be the entropy of normalized probability distribution of $\mathbf{z}_{k}$. 
Intuitively, if $\mathcal{H}_k$ is relatively small, it indicates that the outputs of $g_{cls_k}(\cdot)$ exhibit higher confidence in predicting the relevance. Conversely, a larger $\mathcal{H}_k$ implies a greater degree of uncertainty in the predictions.
We utilize the softmax function to normalize the reciprocals of these entropies:
\begin{equation}
\label{eq:fusion-1}
    \mathbf{w} = \text{softmax} ([1/\mathcal{H}_0, 1/\mathcal{H}_1, 1/\mathcal{H}_3, 1/\mathcal{H}_3]).
\end{equation}

The fused relevance (logits) is obtained through the weighted sum of logits $\{ \mathbf{z}_{0}, \cdots, \mathbf{z}_{3} \}$. And we employ the cross entropy to calculate the matching loss:
\begin{equation}
\label{eq:loss_match}
\begin{split}
    \mathbf{z} &= \sum\nolimits_{k=0}^{3} \mathbf{w}_{k} \mathbf{z}_{k}, \\
    \mathcal{L}_{mat} &= -\sum\nolimits_{j=1}^{\lvert \mathcal{Y} \rvert} \mathbf{y}_{j} \log (\text{softmax}(\mathbf{z})_{j}).
\end{split}
\end{equation}
where $\mathbf{y}$ is the ground-truth label. Finally, the overall loss for DLF-CCM is defined to balance the matching learning and LF learning:
\begin{equation}
\label{eq:loss_total}
    \mathcal{L} = \mathcal{L}_{mat} + \lambda_1 \mathcal{L}_{ex} + \lambda_2 \mathcal{L}_{sh}.
\end{equation}

\section{Experiments}

\subsection{Experimental Setup}
\subsubsection{Dataset} In our experiments, the datasets employed involve a widely used CCM dataset LeCaRD \cite{LeCaRD} and a large-scale LJP dataset CAIL \cite{cail}. All included criminal cases in both datasets were officially published by the Supreme People’s Court of China.

\textbf{LeCaRD} is originally constructed for the criminal case retrieval task, which contains 107 source (query) cases and 43,000 target cases. We follow the data construction and partitioning protocols outlined in \cite{law-match}. Specifically, for each query, 30 target cases are manually annotated, with each case being assigned a 4-level relevance (matching) label. The dataset is divided into training/valid/test subsets with a ratio of 0.8/0.1/0.1, respectively.

\textbf{CAIL} encompasses over 1.68 million criminal cases. It is divided into two sub-datasets of varying sizes. For our study, we employ the larger one, containing approximately 1.58 million cases. Following the preprocessing steps suggested in \cite{ladan}, we have excluded cases with multiple labels as well as with excessively brief textual content.

\subsubsection{Baselines} The following competitive baselines are employed:

\textbf{Sentence-BERT} \cite{sentence-bert} uses siamese network structures of BERT \cite{bert} to encode two sentences separately. Then the two embeddings are concatenated and fed to a classifier to conduct matching.

\textbf{Lawformer} \cite{lawformer} is a Longformer-based \cite{longformer} language model for legal long cases (LLC) understanding, which is pre-trained on a large-scale corpus of Chinese LLC documents. The mean pooling of Lawformer’s output of two cases is used to conduct matching.

\textbf{BERT-PLI} \cite{bert-PLI} segments cases into paragraphs and utilizes BERT to capture the semantic relationships at the paragraph-level of cases; then employs the RNN with attention mechanism to aggregate paragraph-level embeddings of two cases.

\textbf{Law-Match} \cite{law-match}, the SOTA method, is a causal learning framework, which decomposes the treatments (cases) into law article-related and -unrelated parts, and combines them with different weights to collectively support the matching prediction.

\subsubsection{Implementation Details} All methods utilize Legal-BERT\footnote{https://github.com/thunlp/OpenCLaP} as the backbone to encode cases except Lawformer. For the networks $f_{cls_{k}}$ and $g_{cls_{k}}$ in DLF-CCM, we employ a single fully connected layer followed by a softmax activation function (SAF). For the discriminator $D$, we employ three fully connected layers followed by a SAF. For hyperparameters $\lambda_1$ and $\lambda_2$, we set them to $0.05$ and $0.05$.
We employ the Adam optimizer with a learning rate of $2 \times 10^{-5}$ for both training stages. And the (epoch, batch size) for two stages is set to ($10$, $8$) and ($5$, $4$) respectively. Following the work of \cite{law-match}, we employ Accuracy (Acc.), Macro-Precision (MP), Macro-Recall (MR), and Macro-F1 (MF1) as the evaluation metrics.

\subsection{Results}
\begin{table}[!t]
  \caption{Primary results and ablation results.}
  \label{exp:res}
  \begin{tabular}{l|cccc}
    \toprule
     Methods & Acc. ($\%$) & MP ($\%$) & MR ($\%$) & MF1 ($\%$) \\
    \midrule
    Sentence-BERT & 59.44 & 59.54 & 57.89 & 58.70 \\
    Lawformer & 59.13 & 58.79 & 58.56 & 58.47 \\
    BERT-PLI & 61.60 & 60.88 & 60.41 & 60.48 \\
    Law-Match & 65.63 & 66.07 & 63.75 & 64.41 \\

    \midrule
    \textbf{DLF-CCM} & \textbf{68.42} & \textbf{70.39} & \textbf{68.55} & \textbf{68.98} \\

    \midrule
    \midrule
    -~$\mathcal{L}_{ex}$ & 66.56 & 68.33 & 65.47 & 66.54 \\
    -~$\mathcal{L}_{sh}$ & 64.71 & 68.61 & 63.47 & 65.29 \\
    -~Fusion & 64.40 & 67.74 & 62.71 & 64.42 \\
    -~Pre-Train & 63.78 & 63.02 & 63.60 & 62.39 \\

  \bottomrule
\end{tabular}
\end{table}

Table~\ref{exp:res} shows the primary experimental results (top portion) and the associated ablation results (bottom portion). 
(1) We can find that the models (BERT-PLI, Law-Match, DLF-CCM) that are carefully designed for the case matching task perform better than other baselines. This demonstrates a specialized model is essential in this domain. Among baselines, Law-Match shows good performance over other baselines due to the introduction of law articles knowledge. 
We use t-test with significance level 0.05 to test the significance of performance difference, and results show the proposed FLD-CCM significantly outperforms all the baselines on all metrics.
(2) For ablation experiments: -~$\mathcal{L}_{ex}$ or -~$\mathcal{L}_{sh}$ mean we set $\lambda_{1}$ or $\lambda_{2}$ as 0 to demonstrate the effectiveness of LF de-redundancy learning; -~Fusion means that we treat each LF equally in predictions; -~Pre-Train means that we remove the pre-training stage. we can see that the performance of these variants drops apparently, confirming the effectiveness of the four mechanisms.

\begin{figure}[!t]
    \centering
    \includegraphics[width=0.35\textwidth]{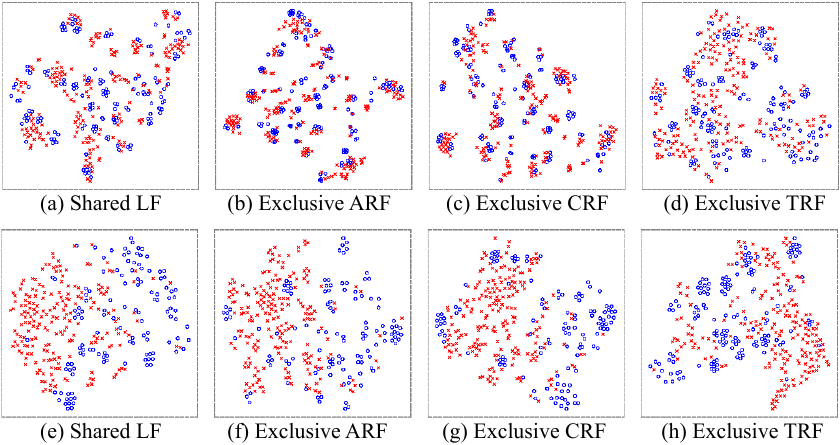}
    
    \caption{The t-SNE plot of LFs. ``Blue'' represents source cases and ``red'' represents target cases. The first and second rows represent case pairs with labels of 3 and 0, respectively.}
    \Description{t-SNE plot of DLF-CCM.}
    \label{exp:vis}
\end{figure}

\subsection{Analysis}

\begin{figure}[!t]
    \centering
    \includegraphics[width=0.4\textwidth,height=0.14\textwidth]{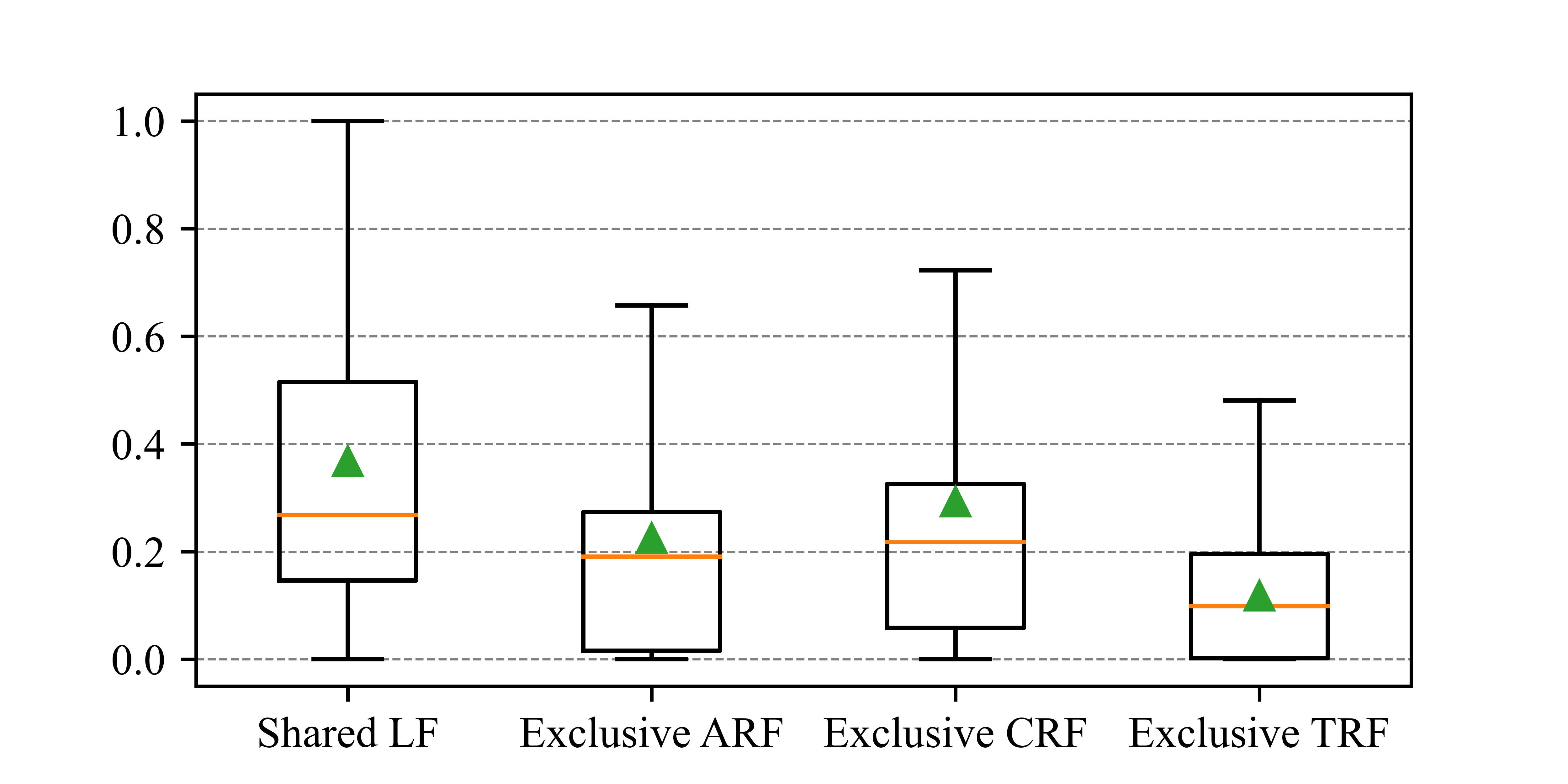}
    
    \caption{The box plot of fusion weights.}
    \Description{.}
    \label{exp:fusion}
\end{figure}

\textbf{Visualization of LFs:} We employ t-SNE to visualize distributions of learned diverse LFs. Specifically, we randomly selected 200 pairs of matched cases (label=3) and 200 pairs of mismatched cases (label=0). The shared LF and exclusive ARF/CRF/TRF of matched cases and mismatched cases are presented in the first row and second row of Figure~\ref{exp:vis}, respectively. 
For matched pairs, we can clearly observe that the LFs of the target cases (red) are mostly distributed around those of the source cases (blue). 
The LFs of unmatched case pairs, in contrast, exhibit two distinct distributions - one for the source cases and one for the target cases.
\textbf{Fusion Weights Analysis:} We further analyze the weights of relevance (logits) associated with different LFs during the relevance fusion process. Figure~\ref{exp:fusion} shows the box plot of these weights. 
Notably, the prediction confidences generated by different LFs vary significantly. The shared LF demonstrates the highest confidence, followed by exclusive CRF. And the exclusive TRF exhibits the lowest confidence, consistent with the findings in Figure~\ref{exp:vis} (where LFs of more target cases in subplots (d) do not consistently center around the source cases.)

\section{Conclusion}

In this paper, we propose a novel method for criminal case matching, DLF-CCM. We design a two-stage framework to effectively learn diverse LFs and remove the redundancy among them. An entropy-weighted fusion strategy is proposed to aggregate the multiple relevance produced by LFs.
Experimental results confirm the effectiveness of DLF-CCM compared to competitive baselines. 
A limitation of our work is that it has achieved good performance only in the task of CCM. For other legal case matching tasks, such as civil cases, we find that defining diverse LFs is challenging and ambiguous. 
This is because court decisions lack diversity and the decisions across cases do not consistently follow the same pattern.
We leave the exploration of this issue to future research.

\begin{acks}
This research was supported by the National Natural Science Foundation of China (Grant Nos. 62133012, 61936006, 62103314, 62073255, 62303366), the Key Research and Development Program of Shaanxi (Program No. 2020ZDLGY04-07), Innovation Capability Support Program of Shaanxi (Program No. 2021TD-05) and Natural Science Basic Research Program of Shaanxi under Grant No.2023-JC-QN-0648.
\end{acks}

\bibliographystyle{ACM-Reference-Format}
\balance
\bibliography{sample-base}

\end{document}